\documentclass[nonacm, sigconf]{acmart}
\renewcommand\footnotetextcopyrightpermission[1]{} 

\def\BibTeX{{\rm B\kern-.05em{\sc i\kern-.025em b}\kern-.08emT\kern-.1667em\lower.7ex\hbox{E}\kern-.125emX}}
    
\usepackage{nicefrac}
\usepackage{siunitx}
\usepackage{array,framed}
\usepackage{booktabs}
\usepackage{
  color,
  float,
  epsfig,
  wrapfig,
  graphics,
  subcaption
}
\usepackage[]{graphicx}
\usepackage{textcomp}
\usepackage{setspace}
\usepackage{latexsym,fancyhdr,url}
\usepackage{enumerate}
\usepackage[ruled,vlined,linesnumbered]{algorithm2e}
\usepackage{algpseudocode}
\usepackage{graphics}
\usepackage{graphicx}
\usepackage{xparse} 
\usepackage{xspace}
\usepackage{multirow}
\usepackage{csvsimple}
\usepackage{balance}
\usepackage{caption}
\usepackage{tabularx}
\usepackage{makecell}
\usepackage{
  tikz,
  pgfplots,
  pgfplotstable
}
\usepackage{hyperref}

\usetikzlibrary{
  shapes.geometric,
  arrows,
  external,
  pgfplots.groupplots,
  matrix
}
\newcommand{\Tau}{\mathrm{T}}
\pgfplotsset{compat=1.9}

\usepackage{mathtools,}

\DeclareMathAlphabet{\mathcal}{OMS}{cmsy}{m}{n}

\DeclareGraphicsExtensions{%
    .png,.PNG,%
    .pdf,.PDF,%
    .jpg,.mps,.jpeg,.jbig2,.jb2,.JPG,.JPEG,.JBIG2,.JB2}

\usepackage{xparse}
\newcommand{\bnm}{\begin{newmath}}
\newcommand{\enm}{\end{newmath}}

\newcommand{\bea}{\begin{eqnarray*}}%
\newcommand{\eea}{\end{eqnarray*}}%

\newcommand{\bne}{\begin{newequation}}
\newcommand{\ene}{\end{newequation}}

\newcommand{\bal}{\begin{newalign}}
\newcommand{\eal}{\end{newalign}}

\newenvironment{newalign}{\begin{align}%
\setlength{\abovedisplayskip}{4pt}%
\setlength{\belowdisplayskip}{4pt}%
\setlength{\abovedisplayshortskip}{6pt}%
\setlength{\belowdisplayshortskip}{6pt} }{\end{align}}

\newenvironment{newmath}{\begin{displaymath}%
\setlength{\abovedisplayskip}{4pt}%
\setlength{\belowdisplayskip}{4pt}%
\setlength{\abovedisplayshortskip}{6pt}%
\setlength{\belowdisplayshortskip}{6pt} }{\end{displaymath}}

\newenvironment{newequation}{\begin{equation}%
\setlength{\abovedisplayskip}{4pt}%
\setlength{\belowdisplayskip}{4pt}%
\setlength{\abovedisplayshortskip}{6pt}%
\setlength{\belowdisplayshortskip}{6pt} }{\end{equation}}

\newcounter{ctr}

\newcounter{mytable}
\def\mytable{\begin{centering}\refstepcounter{mytable}}
\def\endmytable{\end{centering}}

\newcounter{myfig}
\def\myfig{\begin{centering}\refstepcounter{myfig}}
\def\endmyfig{\end{centering}}

\newlength{\saveparindent}
\newlength{\saveparskip}
\newcommand{\E}{{\rm I\kern-.3em E}}

\renewcommand{\eqref}[1]{\mbox{Equation~(\ref{#1})}}

\def \part {part}

\renewcommand{\paragraph}[1]{\vspace*{6pt}\noindent\textbf{#1}\;}

\def \blackslug{\hbox{\hskip 1pt \vrule width 4pt height 8pt
    depth 1.5pt \hskip 1pt}}
\def \qed{\quad\blackslug\lower 8.5pt\null\par}

\newcounter{mynote}[section]

\newcommand\ignore[1]{}

\newcounter{rcnote}[section]

\newcounter{mrnote}[section]

\newcounter{fknote}[section]

\newcounter{anote}[section]

\DeclareMathSymbol{\mlq}{\mathord}{operators}{``}
\DeclareMathSymbol{\mrq}{\mathord}{operators}{`'}

\newcommand{\rhf}[2]{R_{f, \gamma}}

\DeclareDocumentCommand{\edist}{o o}{
  \ensuremath{
    \IfNoValueTF{#1}{{d}}{{\sf d}(#1,#2)}
  }
}

\newcommand{\olrk}[1]{\ifx\nursymbol#1\else\!\!\mskip4.5mu plus 0.5mu\left(\mskip0.5mu plus0.5mu #1\mskip1.5mu plus0.5mu \right)\fi}

\NewDocumentCommand{\indseq}{ O{1} O{r} }{{#1}\ldots {#2}}

\begin{document}

\fancyhead{}
\def\thetitle{Deep Reinforcement Learning for Shared Autonomous Vehicles (SAV) Fleet Management}
\title{\thetitle}

\author{Sergio Sainz-Palacios}
\affiliation{ \href{mailto:ssainz@vt.edu}{ssainz@vt.edu} \\ \small{Virginia Polytechnic}}
\date{}

\begin{abstract}
  Shared Automated Vehicles (SAVs) Fleets companies are starting pilot projects nationwide. In 2020 in Fairfax Virginia it was announced the first Shared Autonomous Vehicle Fleet pilot project in Virginia. SAVs promise to improve quality of life. However, SAVs will also induce some negative externalities by generating excessive vehicle miles traveled (VMT), which leads to more congestions, energy consumption, and emissions. The excessive VMT are primarily generated via empty relocation process.  
Reinforcement Learning based algorithms are being researched as a possible solution to solve some of these problems: most notably minimizing waiting time for riders. But no research using Reinforcement Learning has been made about reducing parking space cost nor reducing empty cruising time.
This study explores different \textbf{Reinforcement Learning approaches and then decide the best approach to help minimize the rider waiting time, parking cost, and empty travel}. 

\end{abstract}

\maketitle
\begin{sloppypar}

\section{Introduction}
\label{sec:intro}

Shared Automated Vehicles (SAV) hold great promise to improve quality of life by providing reliable and affordable door-to-door mobility service to citizens. The technology is advancing rapidly, with many competing companies developing or deploying Automated Vehicle (AV) technology. As of July 20th, 2018, there are 55 Autonomous Vehicle Testing Permit holders who are allowed to test AVs on road in California, increasing from approximately 40 holders in 2017 \cite{californiadmv2018}. Waymo, Google's satellite company in charge of AV, is testing and expanding the fleet of AVs in California and Phoenix, AZ suburbs. Uber, recently resumes its SAV service program in Pittsburgh \cite{economistAV}.  In  2020 in Fairfax, Virginia it was announced the first Shared Autonomous Vehicle Fleet pilot project in the state \cite{fairfax}.

The SAVs are also expected to have positive social impacts \cite{randAV2}, \cite{milakis2017policy}. It is expected that SAV will reduce road accidents that are currently related with human errors (i.e., around 90\% of all accidents). Most of the simulation results show that the system if widely adopted can reduce household vehicle ownership by approximately 90\% \cite{FAGNANT20141}. While the privately owned AVs (PAVs) may only be able to reduce 9.5\% vehicle ownership by sharing AVs among household members \cite{zhang2018impact}. Vehicle ownership reduction can also be translated into significant decline in urban parking demand. Zhang \& Guhathakurta's simulation studies show SAV can reduce the parking land by over 90\% using both grid-based \cite{zhang2015exploring} and empirical travel and road network data \cite{Zhang2017} by reducing vehicle ownership. This indicates that these aesthetically unpleasant and environmentally unfriendly parking spaces can be eliminated and replaced with more sustainable and efficient land uses in an SAV-dominant future. Meanwhile, the SAVs tend to consume less energy and generate Greenhouse Gas (GHG) emissions from the life cycle perspective compared with PAVs and current conventional vehicles \cite{greenblatt2015autonomous,FAGNANT20141}. 


However, there remain concerns regarding the sustainability of SAVs. Simulation studies show the system may generates approximately 10-20\% of the excessive Vehicle Miles Traveled (VMT) compared with private conventional vehicles \cite{FAGNANT20141,Zhang2017}. The extra VMT are typically induced during the vehicle relocation process for passenger pick-up, spatial empty vehicle rebalancing, and parking purposes. One critical step to curb empty VMT generation is to design more efficient vehicle relocation algorithms that can balance the trade-offs between vehicle parking fees, cruising, and the average waiting time of clients. However, to date, only a few studies have explored the design of such algorithm. For instance, Fagnant and Kockelman \cite{FAGNANT20141} examined several different relocation strategies in their simulation model and identified one deterministic algorithm that outperforms the rest as the preferred methods, which was later broaderly adopted in many other simulation studies \cite{Zhang2017, chen2016operations}. In this study, we evaluate whether Artificial Intelligence (AI), specifically the Reinforcement Learning techniques, could improve the vehicle relocation policy by reducing user waiting time, fuel cost and empty VMT generation. As baseline we will use the algorithm developed by Fagnant and Kockelman  \cite{FAGNANT20141}. 
\section{Background and Related Work}

\subsection{Related Work}

Fagnant \& Kockelman (2014) developed a vehicle relocation algorithm for the SAV system \cite{FAGNANT20141}.  The algorithm calculates balancing values for big zones in the hypothetical city to determine the potential relocation destination for idling vehicles located in SAV supply surplus areas. The imbalance value for each zone is estimated based on the following formula:

\begin{equation}
  \text{BlockBalance} = SAV_{S_{\text{Total}}} (\frac{SAV_{S_{\text{Block}}}}{SAV_{S_{\text{Total}}}} - \frac{\text{Demand}_{\text{Block}}}{\text{Demand}_{\text{Total}}})
\end{equation}

\vspace{3mm}
Where, $SAV_{S_{\text{Total}}}$ is the total number of available SAVs in the system; $SAV_{S_{\text{Block}}}$ is the total of available SAVs in the block; $\text{Demand}_{\text{Block}}$ is the expected demand in the block; $\text{Demand}_{\text{Total}}$ is the expected total demand in the study area. 

The limitation of this algorithm includes the arbitrary selections of (1) the spatial configuration of large zone or blocks for imbalance values calculation; (2) random selection of relocation destination after the determination of relocation blocks. Furthermore, although the relocation algorithm account for the overall supply and demand in the system, the relocation decision is not generated on a centralized basis, i.e., the system relocates one vehicle at a time without considering the spatial distribution of other SAVs that will finish service in the near future. These limitations can potential be alleviated or even eliminated by the introduction of Reinforcement Learning (RL) algorithms.

Within the context of transportation, there has been studies related to using RL to control traffic lights \cite{TrafficControlRL} or \cite{trafficlight}. Other studies like the one from Wei, Wang, Yang \cite{taxiRL} use RL techniques to estimate the order in which passengers will be picked up in a shared taxi simulation. Further the study by Lin K. et.al.\cite{SIM2} explores reinforcement learning approach similar to Deep Q-Learning and introduces novel approach to action space to allow for faster training (just use neighboring zones as potential destinations for SAV reallocations). This study is different from Lin K. study in that we explore several reinforcement learning algorithms and not just Deep Q-Learning. A second difference is that this study uses parking cost as a component of the reinforcement learning reward to optimize. Meanwhile work by Lin K does not include parking costs in the reward function.

\subsection{Reinforcement Learning}

Reinforcement Learning is a branch of machine learning dealing with sequential decision making. Given its loose requirements it is applicable to a wide range of applications. It is most useful when a simulation of an agent in a real world system is available and a mathematical models of the agent and system to optimize are lacking. In these cases, in order to obtain satisfactory optimization result, or improve over reward of existing models we can deploy reinforcement learning algorithms. Great attractive of the reinforcement learning approach is that can be used to optimize simulated system without going into details of modeling the system as long as can be encoded as a \textbf{Markov Decision Process}. Using environment instead of system from now on. Several reinforcement learning algorithms warrant convergence to the optimal solution when the number of states in the environment can fit in computer memory. Nevertheless, many real world environments have very large state-space that cannot fit in modern computers. And this is a problem that limits the application of reinforcement learning.
\\
Deep Reinforcement Learning uses function approximators to estimate the value of a state and this has successfully been proven in research settings to optimize reward obtained in games. These experiments demonstrate techniques to overcome the problem of large state-space problem. Nevertheless, there are no converge warranties in deep reinforcement learning. In order to mitigate divergence different techniques have been proposed such as TRPO, Double Deep Q-Learing and others. 
\\
Fleet Management problem is poised to be optimized by Deep Reinforcement Learning (both its original approach as well as its function approximator approach). Rationale for this are the following: 
\begin{enumerate}
    \item Dynamics of taxi ridership demand can be treated as a black box for the reinforcement learning algorithm to figure out.
    \item As we incorporate more info for each city region such as number of ride request, number of drop offs, time of day the totality of the state space may become too large to hold in memory.
\end{enumerate}
In this introduction to reinforcement learning we introduce the following ideas:
\begin{enumerate}
    \item Markov Decision Processes.
    \item Value Iteration and Policy Iteration.
    \item Monte Carlo
    \item Q-Learning
    \item Value function approximators
    \item REINFORCE
    \item Imitation Learning
    \item Monte Carlo Tree Search
    \item Comparison table
\end{enumerate}

In next subsection we discuss different simulators for fleet management and based on them we discuss which reinforcement learning algorithm makes more sense to evaluate.
\subsection{Markov Decision Processes}
In order to understand Markov Decision Processes we can start with Markov Processes. 
\subsubsection{Markov Processes} are Memoryless Random Processes. They are a sequence of random states that hold the \textbf{Markov Assumption}. A state $s_t$ is a \textbf{Markov} state if and only if the next state , $s_{t+1}$, is independent of its previous history $h_t$.
\begin{align}
    p(s_{t+1}| s_t, a_t) &= p(s_{t+1}| h_t, a_t)
\end{align}
Definition of Markov Process:
\begin{enumerate}
    \item S is a finite set of states (s $\in$ S).
    \item P is a dynamics / transition model that specifies $p(s_{t+1} = s' | s_t = s)$
    \item no rewards involved.
\end{enumerate}

\subsubsection{Markov Reward Processes} are equal to Markov Process, but they add a reward function per state $R(s_t)$ and discount factor for the accumulating rewards , $\gamma$. Usually they return the total sum of rewards after reaching the end state : $G = \sum_{i} \gamma^i R(s) $. We also have new concept \textbf{Value} of a state which represents the expected reward obtained from that state and onwards.
\begin{align*}
    V(s) &= R(s_t) + \gamma \sum_{s'} P(s' | s_t) V(s'); \textbf{Expected sum of rewards} \\
    G &= \sum_{i} \gamma^i R(s) ; \textbf{Discounted sum of rewards}\\
\end{align*}
When talking about \textbf{stochastic} markov reward processes, we mean that when a transition happens from a state to a next state, the selection of which actual state will be next state is draw from a probability distribution ($p(s_{t+1} = s' | s_t = s)$). This is different from deterministic markov reward processes where the next state is predefined to the same next state no matter how many times the transition occurs. Hence, in the case of \textbf{stochastic} markov reward process, the $V(s_t) = \mathbb{E}[G_t]$, meanwhile in deterministic markov reward process $V(s_t) = G_t$.

\subsubsection{Markov Decision Processes} are equal to Markov Reward Processes, and add actions :
\begin{enumerate}
    \item Transition model : $P(s_{t+1}=s'| s_{t} = s, a_t = a)$ , a is action. 
    \item Reward model: $R(s_{t+1}=s'| s_{t} = s, a_t = a)$.
    \item Policies: $\pi(a|s) = P(a_t = a | s_t = s)$
\end{enumerate}
With Markov Decision Processes, the expected sum of discounted rewards also depends on the policy (action taken):
\begin{align*}
    V^{\pi} (s) = R(s, \pi(s)) + \gamma \sum P(s'|s,\pi(s)) V^{\pi} (s')
\end{align*}
\subsubsection{Markov Decision Process Control} is about learning optimal policies. While previous definitions help build framework for us to talk and evaluate different algorithms, what we really are interested with is to find policies that optimize certain criteria and that is the Markov Decision Process Control. We will learn about value iteration and policy iteration that are basic policy search algorithms next section.

\subsection{Value Iteration \& Policy Iteration}
\subsubsection{Value iteration} is an algorithm to find the accurate value of expected sum of discounted rewards for any given state , $V(s)$. With an accurate $V(s)$ we can also find best policy.
\begin{algorithm}[]
\SetAlgoLined
 $V_k(.)=0$ \;
 $V_{k+1}(.)=0$ \;
 \While{convergence}{
  \ForEach{$s \in S$}{
    $V_{k+1}(s) = \text{max}_a R(s,a) + \gamma \sum P(s' | s,a) V_k (s')$\;
  }
  $V_{k} = V_{k+1}$ \;
  $V_{k+1}(.) = 0$ \;
 }
 \caption{Value Iteration}
\end{algorithm}

After this, $V(s)$ function can be used to extract best policy:
\begin{align*}
    \pi_{k+1}(s) &= \text{argmax}_a R(s,a) + \gamma \sum P(s' | s,a) V_k(s')
\end{align*}
In value iteration it is expected that the transition probability, $P(s' | s,a)$, and reward function, $R(s,a)$, are known. Also, \textit{convergence} means that the reward vectors did not change more than a small number $\epsilon$ (usually $\epsilon = 0.001$ or less).
\subsubsection{Policy Iteration} is another algorithm that can be used to find the best policy.
\begin{algorithm}[]
\SetAlgoLined
 policy-stable is false\;
 \While{not policy-stable}{
    $V^\pi_k(.)=0$\;
    $V^\pi_{k+1}(.)=0$\;
    \While{convergence of $V^{\pi}$}{
      \ForEach{$s \in S$}{
        $V^\pi_{k+1}(s) = R(s,\pi(s)) + \gamma \sum P(s' | s,\pi(s)) V^\pi_k (s')$\;
      }
      $V_{k} = V_{k+1}$ \;
      $V_{k+1}(.) = 0$ \;
    }
    $b = \pi(s)$\;
    \ForEach{$s \in S$}{
        $\pi(s) = argmax_a R(s, a) + \gamma \sum P(s'| s,a) V^{\pi_i}(s')$\;
    }
    \If{$b = \pi(s)$}{
        policy-stable is true\;
    }
 }
 \caption{Policy Iteration}
\end{algorithm}
Both Policy Iteration and Value Iteration find the best policy and work as long as the dynamics model and reward models are know (R(.), P(.)). They both are guaranteed to converge to global optimal. At the same time these warrants are only for the cases when state-space can fit in computer memory (aka tabular case). Depending on the environment sometimes policy iteration will be faster to converge, other times will be value iteration so is worth trying both.

\subsection{Monte Carlo Policy Control}
Policy Control involves nuanced challenges mainly: 
\begin{enumerate}
    \item Optimization: maximize the rewards
    \item Delayed reward or credit assignment: this refer to the fact that an action taken at this point will have benefits many time steps from now. How can we keep track of this consequence or credit?
    \item Exploration: given that the environments' states are not immediately visible to the agent, agent needs to explore the environment in order to discover rewards.
\end{enumerate}

Also, there are two categories we can classify control algorithms by:
\begin{enumerate}
    \item On-policy :  learn about a policy by following that policy.
    \item Off-policy :  learn about a policy by following a different policy.
\end{enumerate}
Ideally we want to use off-policy algorithms as they promise to reuse records of other policies' episodes. Nevertheless, due to different policies will have each different action distribution some importance sampling needs to occur which makes the learning less efficient.

\subsubsection{First-visit Monte Carlo On Policy Evaluation} does not require reward model nor transition model and does not even require Markov state assumption.  The main idea is to deduce $V(s) = \frac{1}{n} \sum^n_{i=1} G_i$. Some limitation is that can only be applied to episodic environments.

\begin{algorithm}[]
\SetAlgoLined
 \While{True}{
    $N(s) = 0 , G(s) = 0$ \;
    Sample trajectory$ \Tau= s,a,s_2,a_2,...,s_n$ using policy $\pi$ \;
    \ForEach{$s \in \Tau$}{
        \If{first time that state $s$ is visited in episode}{
            Increment counter of visits $N(s) += 1$ \;
            Increment total return $G(s) = G(s) + G_{i,t}$ \;
            Update estimate $V^\pi (s) = G(s) / N(s)$ \;
        }
    }
 }
 \caption{First Visit Monte Carlo}
\end{algorithm}

\subsubsection{Every-visit Monte Carlo on Policy Evaluation} is same as First-visit MC , but instead of increasing on the first state appearance, it does it for every time. This results in First-Visit MC in being unbiased, meanwhile Every-visit MC in being biased (because is biased for states that appear multiple  times in an episode compared with other states that appear only once in an episode).

\subsection{Q-Learning}
\subsubsection{Q-function} is the estimated discounted sum of rewards for a action/state pair. Similar to $V$ function, but with addition of having action as input. In similar fashion once we know a good estimate of the Q value, we can create optimal policy that picks the action with the highest Q value. 
\begin{align*}
    Q(s,a) =  \sum_{s'} P(s'|s,a) (R(s'|s,a) +  \gamma V(s'))
\end{align*}
\subsubsection{Temporal difference} refers to a trick used to estimate the next state's maximal optimal Q value and with this estimate we can improve upon the Q value of the current state-pair:
\begin{align*}
    Q(s_t,a_t) &\leftarrow Q(s_t,a_t) + \eta ( [r_t + \gamma * max_a Q(s_{t+1},a)] - Q(s_t,a_t) ) \\
\end{align*}
The intuition behind this estimation is that the actual $Q(s_t,a_t)$ value should be the next state's maximum Q value ($max_a Q(s_{t+1},a)$) plus the reward obtained in current step t ($r_t$).

\begin{algorithm}[]
\SetAlgoLined
 \While{True}{
    $Q(.,.)= 0, s_t = s_0, \pi_b (s_t) \text{is } \epsilon  \text{ greedy}$ \;
    Sample trajectory $ \Tau = s,a,s_2,a_2,...,s_n$ using policy $\pi$ \;
    \While{True}{
        $a_t \sim \pi_b (s_t)$ \;
        Observe $r_t, s_{t+1}$ from ENV. \;
        $Q(s_t,a_t) \leftarrow Q(s_t,a_t) + \eta ( [r_t + \gamma * max_a Q(s_{t+1},a)] - Q(s_t,a_t) )$ \;
        $\pi(s_t) \leftarrow argmax_a Q(s_t,a) $ \;
        $t \leftarrow t+1 $ \;
    }
 }
 \caption{Q-Learning}
\end{algorithm}

\subsection{Double Q-Learning and Maximization Bias}

If we take unbiased estimators (e.g. from monte carlo estimation of Q-values or even biased estimators such as Q-Learning), once we update the policy, the policy may become biased, because maybe the number of samples of an state (say for Q function) is too small to reach the actual Q value for that state-action and hence it may be a little higher than the actual and such state-action will be chosen over other state-actions pairs. This is known as \textbf{Maximization Bias}.
\\
One way to solve this \textbf{Maximization Bias} for Q-Learning is to split the Q-function: (1) for the estimation of Q and the other, (2) for pick decision of Q. And this approach is named \textbf{Double Q-Learning}.
 \\
 \begin{algorithm}[]
\SetAlgoLined
 \While{True}{
    $Q_1(.,.)= 0, Q_2(.,.)=0,s_t = s_0, \pi_b (s_t) \text{is } \epsilon  \text{ greedy}$ \;
    Sample trajectory  $\Tau =s,a,s_2,a_2,...,s_n$ using policy $\pi$ \;
    \While{True}{
        $\pi(s_t) \leftarrow argmax_a ( Q_1(s_t,a) + Q_2(s_t,a)  ) $ \;
        $a_t \sim \pi_b (s_t)$ \;
        Observe $r_t, s_{t+1}$ from ENV. \;
        \eIf{With 0.5 probability}{
        $Q_1(s_t,a_t) \leftarrow Q_1(s_t,a_t) + \eta ( [r_t + \gamma * max_a Q_1(s_{t+1},a)] - Q_1(s_t,a_t) )$ \;
        }{
        $Q_2(s_t,a_t) \leftarrow Q_2(s_t,a_t) + \eta ( [r_t + \gamma * max_a Q_2(s_{t+1},a)] - Q_2(s_t,a_t) )$ \;
        }
        $t \leftarrow t+1 $ \;
    }
 }
 \caption{Double Q-Learning}
\end{algorithm}
 \\
 
 \subsection{Value Function Approximators}
 As state-space grows, the previous algorithms cannot be solved as they require for the totality of state-space to be iterated over. Solution for this challenge has been to use Value Function Approximators (VFA) to approximate V function or Q function \cite{ATARI}.
 
 Main concern of using value function approximators is that they break assumption made by previous algorithms that in each iteration of the algorithm (be it value-iteration/policy-iteration/Q-Learning/MonteCarlo) the difference in the V/Q function will be monotonically decreasing. In other words, meanwhile tabular methods the incremental improvement is a contractor, in function approximators the incremental improvement is a contractor as well as an expansor sometimes. \textbf{Because VFA  sometimes are an expansor, then there is no convergence warranties}.
 \\

\subsection{REINFORCE}
 REINFORCE is an algorithm part of a wider family of reinforcement learning algorithms known as Policy gradient algorithms.
 \subsubsection{Policy Gradient} algorithms do not attempt to model the discounted sum of rewards until end of episode (either V or Q) and instead just focus on modifying the policy $\pi(s) = a$ in such a way as to increase the reward. This is usually done by using Artificial Neural Network to approximate the policy and applying stochastic gradient descent algorithm on sum of discounted rewards , $J(\theta)$ w.r.t to the policy function approximation weights $\theta$:
 
\begin{align*}
    J(\theta) &= V(\theta) = \sum_i \gamma^i r_i \\
    \nabla_\theta J(\theta) &= \mathbb{E}[\nabla_\theta log \pi_\theta (s,a) * Q^{\pi_\theta}(s,a)]
\end{align*}

Here are just the results, please check \cite{Sutton} for the derivation. Using above gradient we can do the following idea: Sample a trajectory, then apply gradient descent on each step of that trajectory and then repeat. That in essence is REINFORCEMENT algorithm,

\subsubsection{REINFORCE} is as below:

\begin{algorithm}[]
\SetAlgoLined
Init $\theta$ randomly \;
 \While{K number of times}{
    Sample trajectory $\Tau = s,a,s_2,a_2,...,s_n$ using policy $\pi$ \;
    \ForEach{$s_i,a_i,r_i in \Tau$}{
        $G_t \leftarrow \sum^{j=n}_{j=i+1} \gamma^j r_j$ \;
        $\theta \leftarrow \theta + \alpha \nabla_\theta log \pi_\theta (s_i, a_i) G_t$ \;
    }
 }
 \caption{REINFORCE}
\end{algorithm}

 \textbf{Advantages} of using policy gradient algorithm like REINFORCE are:
\begin{enumerate}
    \item Sometimes is easier to encode policy rather than value functions (V,Q).
    \item allows to learn stochastic policies.
    \item Better convergence properties.
\end{enumerate}
\textbf{Disadvantages}:
\begin{enumerate}
    \item Typically evaluate to local maximum rather than global optimum. So algorithm needs to be run several times with neural network weights initiated in different random states and just remember the best performing ending neural network weights.
    \item Evaluating policy is inefficient and high variance.
\end{enumerate}

\subsection{Imitation Learning}
Two big challenges of reinforcement learning: credit assignment problem and exploration are solved if we demonstrate reinforcement learning agent the behavior we wish it to improve on. In Imitation Learning we encode domain knowledge in the form of policy class used. That is, instead of learning from scratch we offer reinforcement learning agent another policy to imitate. This helps bootstrap the learning and afterwards agent has learned the bootraped policy it can continue with independent learning.
\\
In imitation learning, we try to produce the same discounted state distribution between the teacher policy and the student policy and some research suggests on using Generative Adversarial Networks to do this.
\\
\subsection{Monte Carlo Tree Search (MCTS)}
MCTS uses model-based learning to simulate state, action, state trajectories and with them select the action that is the most promising after n number of simulations. It is often uses in game board environments \cite{GOGAME}.

\subsubsection{Model-based learning} is often times useful when the model is easy to learn. Then, we can use value iteration / policy iteration to find best policy. Or in case the state-space is too large, we can use other value function approximators reinforcement learning algorithms to learn from learnt model (that can produce many data points relatively cheaply compared with the real world).
\\
By model we refer to the transition function of state, action to another state. And to the reward function given a state and action.
\\
\begin{itemize}
    \item Model $ M = {P, R}$, transition and reward functions.
    \item $S_{t+1} \sim P(S_{t+1}| S_t, A_t)$
    \item $R_{t+1} \sim R(R_{t+1}| S_t, A_t)$
    \item assume conditional independence between rewards and state transitions: $\mathbb{P}[S_{t+1},R_{t+1} | S_t, A_t] = \mathbb{P}[S_{t+1}| S_t, A_t]\mathbb{P}[R_{t+1}| S_t,A_t]$
\end{itemize}
Goal is to learn model based on experiences, is a supervised learning problem (features are the previous state and action and target are the reward and state):
\begin{align*}
    S_1, A_1 &\rightarrow R_2,S_2 \\
    S_2, A_2 &\rightarrow R_3,S_3 \\
    S_3, A_3 &\rightarrow R_4,S_4 \\
    ...&\\
    S_{T-1}, A_{T-1} &\rightarrow R_T,S_T \\
\end{align*}
Learning $s,a \rightarrow r$ is a regression learning. Learning $s,a \rightarrow s'$ is a density estimation problem. Pick loss function, hyperparameters.
\subsubsection{Monte Carlo Tree Search} algorithm can be described as simulating several trajectories that are most promising based on previous trajectories results for as long as possible and once time to take decision comes, picks the best next action based on previous simulations. While MCTS is simulating trajectories they are recorded in the tree search. States are nodes in the tree. As states in the tree are revisited, their value functions are recalculated for future use. 
\begin{enumerate}
    \item Precondition is to have an existing model.
    \item Start with empty tree in state zero.
    \item Simulate a trajectory and record it the tree.
    \item As new trajectory is about to start, select next action based on the action that renders maximum Q function : $Q(s,a,i) = \frac{1}{N(i)}\sum_{k=1}^K \sum_{u=t}^{T} (i \in \text{epi.k}) G_k(s,a,i) + c\sqrt{\frac{ln(n(s))}{n(s,a)}}$.
    \item $i$ is the ith episode, 
    \item $N(i)$ is the number of times the episode up to this state has been visited. 
    \item in general this Q function reuses part of the upper confidence bounds and normal monte carlo search to select the best action to visit next. Such that the action with the most confidence and highest value is selected.
\end{enumerate}
Advantages of MC tree search
\begin{itemize}
    \item Highly selective best-first search. (does not need to build all the state-space tree).
    \item Evaluates states dynamically (like Dynamic Programming).
    \item Uses sampling to break curse of dimensionality.
    \item Works for black-box models.
    \item Computational efficient and parallelisable.
\end{itemize}
Disadvantages of MC tree search
\begin{itemize}
    \item Could be that one branch of the tree has a single loss action but is pruned by the tree and hence not explored.
    \item Requires a model to be available, then limited use to those environments with simple model or to environments whose model are available or can be learned.
\end{itemize}

\subsection{Comparison Table}
Please refer to table \ref{comp}.
\begin{table*}[t]
\centering
\begin{tabular}{|l|l|l|l|l|l|l|}
\hline
Algorithm \textbackslash Metric    & Requires Markov & Model based  & Biased or unbiased                          & Supports            & Convergence                  & Supports Large           \\
    & Assumption & or model free & Biased or unbiased                          & Stochastic Policies           &  Warranties                 &  State-Spaces           \\
\hline
Value Iteration / Policy Iteration & Yes                       & Model based               & Unbiased                                    & No                                     & Yes                                    & No                                    \\ \hline
Monte Carlo First Visit            & No                        & Model free                & Unbiased                                    & Yes                                    & Yes                                    & No                                    \\ \hline
Monte Carlo Every Visit            & No                        & Model free                & Biased+ & Yes                                    & Yes                                    & No                                    \\ \hline
Q-Learning                         & Yes                       & Model free                & Biased+ & No                                     & Yes                                    & No                                    \\ \hline
Deep Q-Learning                    & Yes                       & Model free                & Biased+ & No                                     & No                                     & Yes                                   \\ \hline
Double Deep Q-Learning                    & Yes                       & Model free                & Biased+ & No                                     & No                                     & Yes                                   \\ \hline
REINFORCE                          & Yes                       & Model free                & Biased+ & Yes                                    & No                                     & Yes                                   \\ \hline
Imitation Learning                 & Yes                       & Model free                & Depends*    & Depends* & Depends* & Depends* \\ \hline
Monte Carlo Tree Search            & No                        & Model based               & Biased+ & Yes                                    & No                                     & Yes                                   \\ \hline
\end{tabular}
\caption{Comparison of reinforcement learning algorithms. Biased+: means that are biased due to some states being visited more often than others. Depends*: means that depending on underlying function approximator implementing the  teacher's policy }
\label{comp}
\end{table*}

\section{Methodology}

\subsection{Simulators}
There are several simulators we can use and the criteria to choose one is based on the following:

\begin{enumerate}
    \item able to provide feedback about travel time, customer's waiting time as well as parking costs? 
    \item How easy is it to adapt to reinforcement learning loop of execute action, receive next state, reward, then take another action?
    \item And more technical details is whether it exists in python as reinforcement learning algorithms author has are coded in python. 
\end{enumerate}
\subsubsection{Simulator I: Exploring the Impact of Shared Autonomous Vehicles on Urban Parking Demand: An Agent-based Simulation Approach} paper by W. Zhang et.al.  \cite{ZHANG} developed a simulator that allows to model waiting times, travel times, and parking costs. More over, it is based upon Atlanta City traffic patters and areas. This is the most sophisticated environment. At the same time, simulator does not fit nicely into the reinforcement learning framework because time steps are not discrete but rather continuous. This means there is no clear cut place where we can measure the reward obtained by the agent. Still this problem could be mitigated by adding a step once the agent finishes executing the action.

\subsubsection{Simulator II: Efficient Large-Scale Fleet Management via Multi-Agent Deep Reinforcement Learning} paper by Lin K. et.al.\cite{SIM2} developed another simulator that allows to model waiting times and travel times. It is discrete time based and hence it allows for easier integration with Reinforcement Learning training process. It is based on square or hexagonal synthetic areas while actual traffic data can be loaded upon it.  Disadvantage of this simulator is that does not have parking time information modeled into the simulation.

\subsubsection{Simulator I as selected testing platform}
The cost associated with adjusting the workflow of simulator I to make its time-step discrete compared with the cost of adjusting the workflow of simulator II to add parking lot costs causes us to decide on simulator I. Especially because we lack data to augment simulator II with parking lots cost. 

\subsubsection{Simulator workflow}

\textit{Simulator I} follows OpenAI's gym Application Programmable Interface (API) \cite{gym} for simulators:

\begin{enumerate}
    \item \textbf{reset()} : restarts the environment to a new episode and returns the initial state.
    \item \textbf{step(a)} : takes action a in environment and returns the following tuple: (state, reward, is episode finished? , extra info)
    \item \textbf{close()} : deletes environment's internal data structures.
\end{enumerate}

And then we can easily adjust reinforcement learning loop of : observe state, take action on environment, get reward , observer new state, take new action, get new reward, etc , using a sequence of calls to the \textbf{reset()}, \textbf{step($a_1$)}, \textbf{step($a_2$)}, ..., \textbf{step($a_n$)}, \textbf{reset()} APIs. 
\\
The meaning of action (\textbf{a}) parameter in the \textbf{step(a)} API is interpreted like this: \textit{Each time an agent is free to pick up for customers, the vector a is sampled to identify which zone to travel to}. The expectation is that the agent will learn to identify how to best allocate SAV in city's zones as the demand changes according to the hour of the day.

\textbf{Note} that in this simulator each episode lasts for 7 days and there is one step every hour (for a total of 24*7 steps per episode). During step-hour the decision made by the agent is kept constant until next step-hour arrives and a new decision is made. 

The decision of using 7 days to represent one episode is arbitrary and kept to represent one week time. But there is no reason to believe results will be different if using 1 day or 10 days per episode.

\subsubsection{Environment State Encoding}

\textbf{Given a city with $n$ zones}, the environment state is defined by three parts:

\begin{enumerate}
    \item \textbf{Travel demand per zone}:  $\frac{\text{demand}_i}{\sum_{j=0}^{n} \text{demand}_j}$. Where $\text{demand}_i$ is the demand in $ith$ zone.
    \item \textbf{Share of SAV per zone}:  $\frac{\text{SAV}_i}{\sum_{j=0}^{n} \text{SAV}_j}$. Where $\text{SAV}_i$ is the number of SAV in $ith$ zone (busy or otherwise).
    \item \textbf{Parking space per zone}:  $\text{parking}_i$. Where $\text{parking}_i$ is the number of available parking spaces in $ith$ zone.
\end{enumerate}

Then these three vectors of $n$ zones each are concatenated to form a $3n$ one-dimension vector. This is the vector the agent observes as \textit{environment state}.

\subsubsection{Action space}

\textbf{Given a city with $n$ zones}, two approaches to encode the action space are studied:

\begin{itemize}
    \item $n$-zone action space: one action per each one of the zones in the environment. Action space size of $n$. This basically means that for each state agent is at there will be $n$ number of possible actions to choose from.
    \item 4-nearest zones action space: One action per each one of the four closest zones. Action space size of 4. This basically means that for each state agent is at there will be 4 possible actions to choose from.
\end{itemize}

The larger the action space the longer it takes for the value approximators to reach the real Q-value, hence it is expected the 4-nearest zones action space to outperform the $n$-zone action space.

\subsubsection{Computing reward per step}

The reward function incorporates three main costs: customer waiting time, SAV empty travel cost and parking costs. Our algorithm is modeled as a single agent reinforcement learning even though it handles several SAV at the same time. Our proposed RL algorithm models the behavior of a centralized route planner rather than the individual behavior of each SAV.  Because we are modeling the behavior of the central planner, therefore, the reward we are trying to optimize should comprise the rewards for all the SAVs. To achieve this goal we use the average reward of the whole system as the reward. We call it global reward, $R_g$. The global reward depends on the state of the environment at a given time.  The notations are displayed in table \ref{table:rewardcomponents}. Based on that notation the reward is defined as: 

\begin{table}
\centering
\begin{tabular}{ |l|c| } 
 \hline
 Term & Meaning \\ 
 \hline
 $W$ & \thead[l]{set of all clients waiting (max 1 day waiting)}  \\
 \hline
 $W_i$ & \thead[l]{the ith client from set W} \\ 
 \hline
 $W^c_i$ & \thead[l]{the ith client's salary} \\ 
 \hline
 $W^{\text{wa}}_i$ & \thead[l]{the ith client's waiting time} \\ 
 \hline
 $ET$ & \thead[l]{set of all empty vehicles} \\ 
 \hline
 $ET_j$ & \thead[l]{the jth SAV from set $ET$} \\ 
 \hline
 $ET^d_j$ & \thead[l]{the distance jth SAV traveled while empty} \\ 
 \hline
 $P$ & \thead[l]{set of all vehicles parked} \\ 
 \hline
 $P_z$ & \thead[l]{the zth SAV from set $P$ } \\ 
 \hline
 $P^c_z$ & \thead[l]{the parking fee in the \\ location where the zth SAV was parked at}\\ 
 \hline
 $gas$ & \thead[l]{constant price of gas }\\ 
 \hline
\end{tabular}
\caption{Reward components notation}
\label{table:rewardcomponents}
\end{table}

\begin{equation}
R_g(time) = - (\frac{1}{|W|}\sum_i W^c_i * W^{\text{wa}}_i + \frac{1}{|ET|}\sum_j ET^d_i * gas + \frac{1}{|P|}\sum_z P^c_z) \label{eq:reward_global}
\end{equation}

\textbf{Note} the additional negative sign after adding up the three components of the reward, this is because all the components are actually costs and so we want to reduce them rather than encourage them. All RL algorithms will try to maximize the reward and hence reduce the cost (as long as the costs are negative).

\subsection{Baseline: Imbalance Algorithm}

First put forward by Fagnant and Kockelman \cite{FAGNANT20141} , the imbalance algorithm works by moving SAVs from areas with high SAV density to areas with low SAV density. Specifically: given the state from section \textit{Environment State Encoding}, it is comprised by three parts: (1) travel demand, (2) share of SAV per zone, and (3) parking space per zone.
\\
\textbf{imbalance baseline algorithm} focuses in the first two parts:

\begin{enumerate}
    \item \textbf{Travel demand per zone}:  $\frac{\text{demand}_i}{\sum_{j=0}^{n} \text{demand}_j}$. Where $\text{demand}_i$ is the demand in $ith$ zone.
    \item \textbf{Share of SAV per zone}:  $\frac{\text{SAV}_i}{\sum_{j=0}^{n} \text{SAV}_j}$. Where $\text{SAV}_i$ is the number of SAV in $ith$ zone (busy or otherwise).
\end{enumerate}
    
And combines them both into a single $n$ vector (where $n$ is the number of zones):
\\
Each $ith$ element of this vector corresponds to $ith$ zone and is defined by:
\\
\begin{align*}
    \text{imbalance}_i = \frac{\frac{\text{SAV}_i}{\sum_{j=0}^{n} \text{SAV}_j} - \frac{\text{demand}_i}{\sum_{j=0}^{n} \text{demand}_j}}{\frac{\text{demand}_i}{\sum_{j=0}^{n} \text{demand}_j}}
\end{align*}

Afterwards a one-hot encoding $n$ elements vector is build where the $ith$ element is 1 if the $\text{imbalance}_i \ge 10\%$ and 0 otherwise.
\\
Finally the agent samples one element from the distribution specified by one-hot vector, that is, only those elements with imbalance greater than 10$\%$ will be selected when the agent is sampling.

\label{sec:methodology}

\section{Evaluation}
\label{sec:eval}

\subsection{Reinforcement Learning Algorithms to be studied}
The idea of the study is to identify proper reinforcement learning algorithms for the Shared Autonomous Vehicles (SAV) Fleet Management problem in large action states. The following criteria is used to select reinforcement learning algorithms:
\begin{itemize}
    \item Works with large state action spaces. Although the simulator's settings are using only 21 zones (for a 21*3 state space), we want to investigate algorithms that will support large state spaces for future work.
    \item Model-free learning algorithms are preferred as they do not require extra step to create a model. 
\end{itemize}

Thus, the algorithms that comply with these requirements are Deep Q-Learning, Double Deep Q-Learning and REINFORCE. And they will be used in the study.

\subsection{Experiments}

\subsubsection{All-zones action space vs 4-nearest zones action space}
There are two options of encoding the \\ action space as explained in section \textit{Action space}:
\begin{itemize}
    \item \textit{n-zone action space}: in this option there are as many actions as zones and hence it can grow rapidly as we apply the algorithm to larger cities or areas. Usually reinforcement learning algorithms do not perform well on large action spaces  \cite{LargeActionSpaces}. Hence we expect the learning in this action space approach is not going to happen or happen too slowly.
    \item \textit{4-nearest zones action space}: This approach tries to reduce the action space to only the four nearest zones per each zone and hence reduces the action space considerably. Inspiration is taken from work by Lin, Renyu, et.al \cite{Kaixiang} Expectation is that this action space approach will allow rewards to be tractable: as there are only four options to test, the number of example action test in the simulator for a given state is smaller than the \textit{n-zone action space} case. Also, at the time of training, there are only four potential actions to learn from compared with \textit{n-zone action space} that has n actions to learn from. Thus learning will also be faster.
\end{itemize}

\subsubsection{Imbalance algorithm vs Reinforcement Learning algorithms}

We want to measure reinforcement learning algorithms against proven industry algorithms. We use the \textit{Imbalance Algorithm} from Fagnant and Kockelman \cite{FAGNANT20141} and explained in section \textit{Baseline: Imbalance Algorithm}. 
\\
In this experiment we will compare Imbalance algorithm against the three selected RL algorithms : DQN, DDQN and REINFORCE.

\subsection{Experiments setup}

The experiments are executed as follows: Each agent is trained for 7 simulator days (one week) and that is considered one iteration. The reward is obtained every hour of the iteration and such reward is used for training the underlying Reinforcement Learning algorithm. And at the end of the iteration, the iteration reward is computed as the average of the hourly rewards during that iteration. $\textit{iteration reward}_{\text{1st}} = \sum_{i=1}^{168} r_i$. Then, the agent is trained for 100 iterations (100 weeks). Finally the same experiment is repeated again with the same agent starting from random network initialization. And the final reported reward for a given agent for a given iteration is the maximum of both experiments for that iteration. $\textit{iteration reward}_{\text{final}} = max(\sum_{i=1}^{168} r^1_i, \sum_{i=1}^{168} r^2_i)$. Where $r^1_i$ is the reward at ith hour in the first experiment and $r^2_i$ is the reward at ith hour in the second experiment. For example for agent RobotDDQN, for 10th iteration, the value of the reported reward is the maximum between the first experiment's 10th iteration reward and the second experiment's 10th iteration reward.

\subsection{Function Approximator configuration}

A neural network is used as function approximator for all agents and for all agents the architecture is as follows: 
\begin{enumerate}
    \item \textbf{1st layer}: 21*3 neurons
    \item \textbf{2nd layer}: 120 neurons
    \item \textbf{3rd layer}: 120 neurons
    \item \textbf{4th layer}: 21 * n in case of \textit{n-zone action space}, or 21 * 4 in case of \textit{4 nearest zones action space}.
\end{enumerate}

The non-linear function ReLu is applied between each of the layers.
\begin{align*}
    \textit{ReLu}(x) =  max(x,0)
\end{align*}

A softmax function is applied after last layer to ensure the final vector sums up to 1 as the agents will sample from this final action distribution or take its max.

\begin{align*}
    \textit{Softmax}(\mathbb{X}) = \frac{e^{x_i}}{\sum_{j=1}^n e^{x_j}}
\end{align*}

\subsection{Results}

\subsubsection{n-zone action space vs 4-nearest zones action space}

\begin{figure}
  \includegraphics[width=\linewidth]{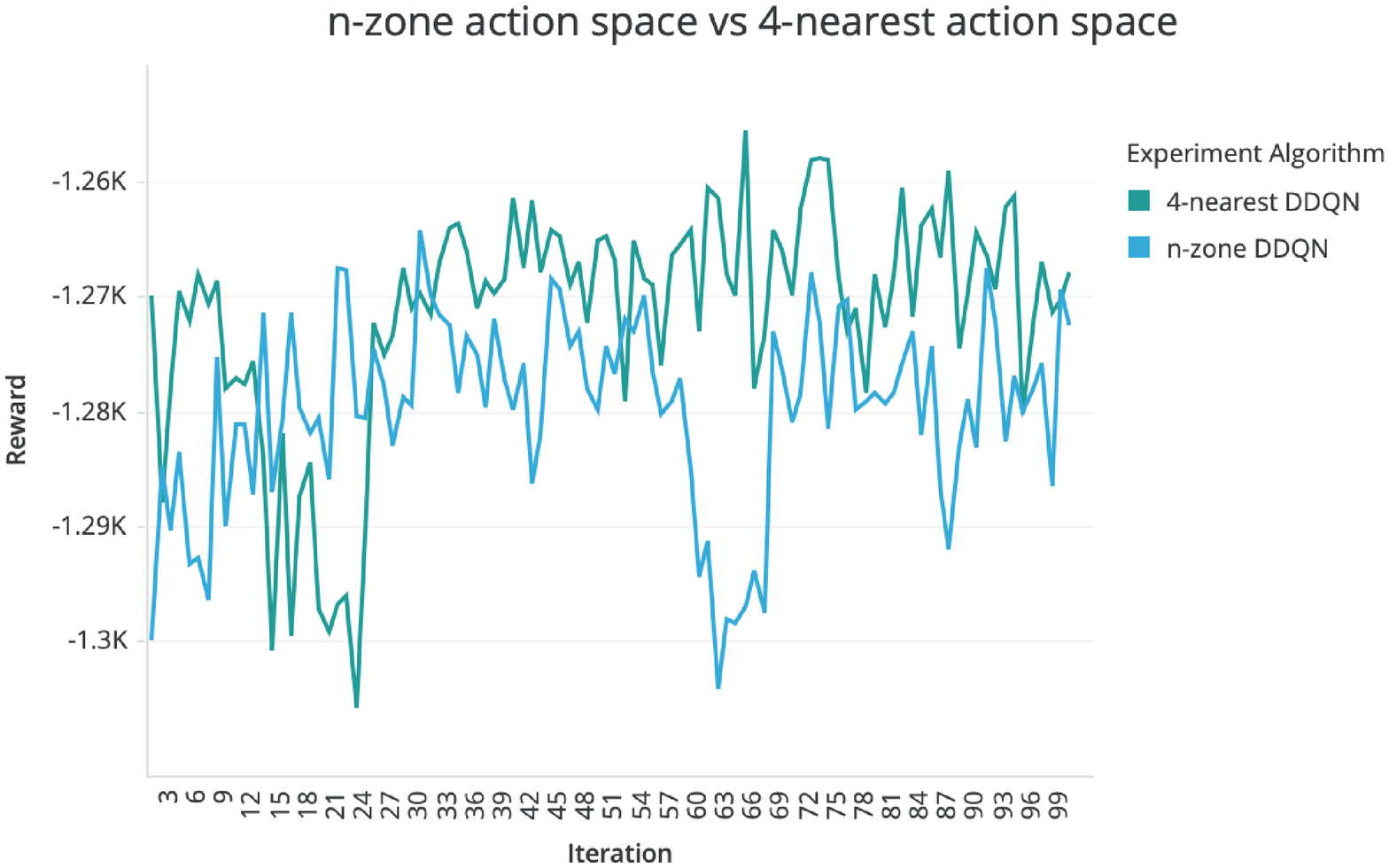}
  \caption{DDQN, n-zone vs 4-nearest action space}
  \label{fig:exp1_ddqn}
\end{figure}

\begin{figure}
  \includegraphics[width=\linewidth]{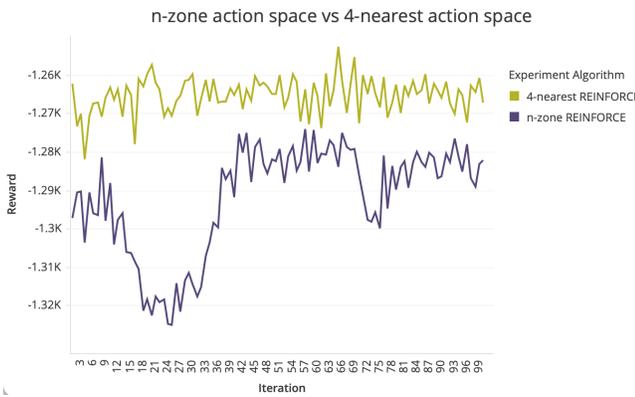}
  \caption{REINFORCE, n-zone vs 4-nearest action space}
  \label{fig:exp1_reinforce}
\end{figure}

The results of comparing \textit{n-zone action space} vs \textit{4-nearest zones action space} is the latter performs better in all three compared algorithms: Deep Q-Learning (DQN), Double Deep Q-Learning (DDQN), and REINFORCE. This matches with the literature \cite{LargeActionSpaces} as well as with expectation that given few possible actions makes it easier to explore the environment as well as easier to backpropagate the reward signal from fewer actions. In DQN case we see in figure \ref{fig:exp1_ddqn} the charted comparison of both approaches and in the summary table that the 4-nearest zones approach outperforms n-zones approach by 1.7 \%.  In DDQN case we can see in figure \ref{fig:exp1_dqn} the charted comparison and the 4-nearest zones outperforms n-zones approach by 0.4 \%. REINFORCE case is in figure \ref{fig:exp1_reinforce} and 4-nearest zones outperforms n-zones approach by 1.5 \% .

\begin{figure}
  \includegraphics[width=\linewidth]{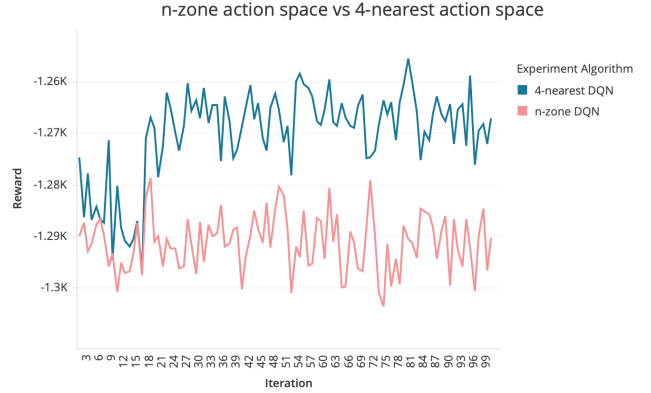}
  \caption{DQN, n-zone vs 4-nearest action space}
  \label{fig:exp1_dqn}
\end{figure}
\begin{figure}
  \includegraphics[width=\linewidth]{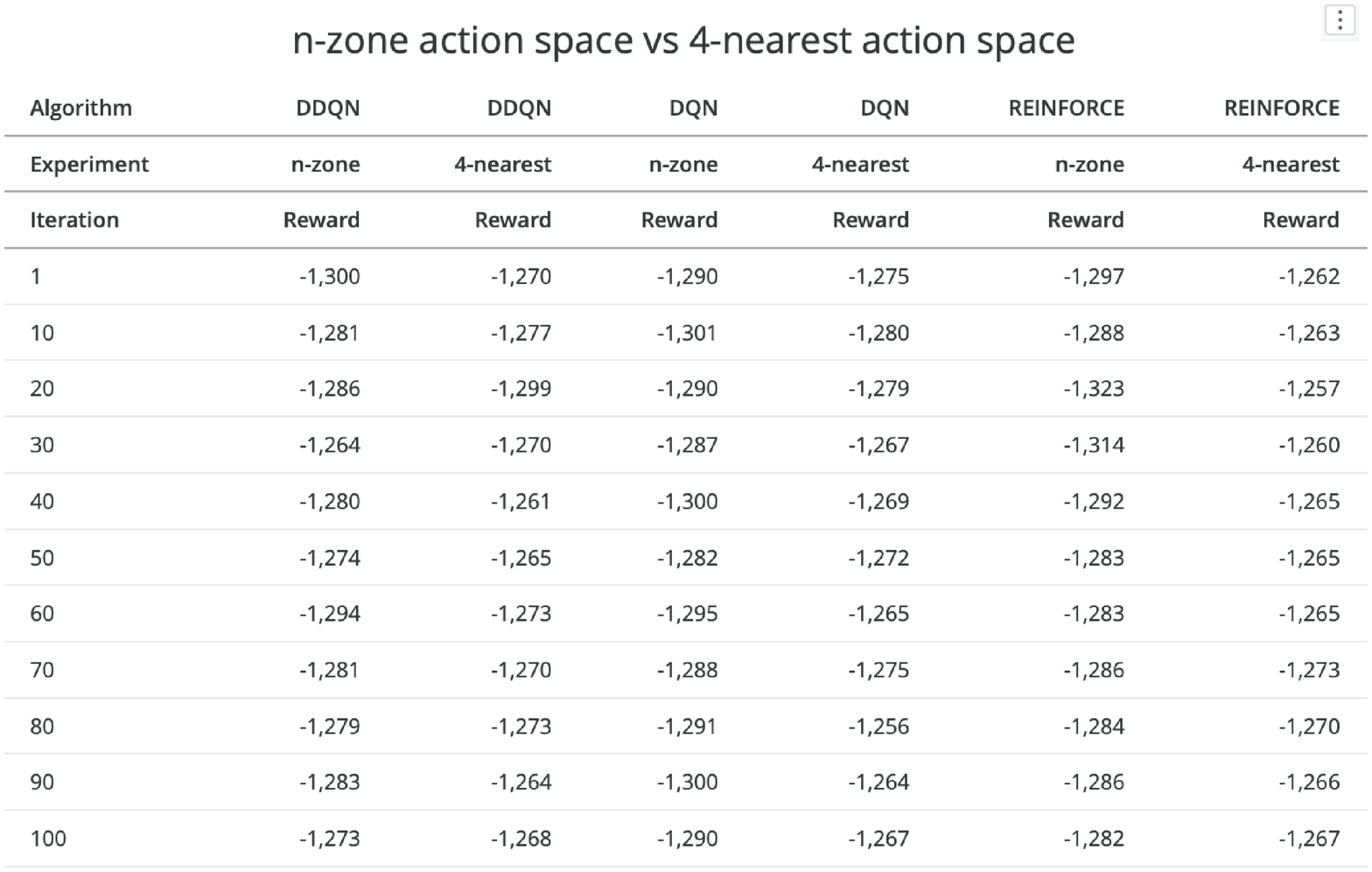}
  \caption{Tabular data, n-zone vs 4-nearest action space}
  \label{fig:exp1_dqn}
\end{figure}

\begin{table}[H]
\begin{tabular}{|l|l|l|}
\hline
Algorithm                  & Action space       & \textit{100th} iteration reward \\ \hline
\multirow{3}{*}{DDQN}      & n-zone             & -1,273                             \\ \cline{2-3} 
                           & 4-nearest neighbor & -1,268                             \\ \cline{2-3} 
                           & \% Improvement     & 0.4\%                              \\ \hline
\multirow{3}{*}{DQN}       & n-zone             & -1,290                             \\ \cline{2-3} 
                           & 4-nearest neighbor & -1,267                             \\ \cline{2-3} 
                           & \% Improvement     & 1.7 \%                             \\ \hline
\multirow{3}{*}{REINFORCE} & n-zone             & -1,287                             \\ \cline{2-3} 
                           & 4-nearest neighbor & -1,267                             \\ \cline{2-3} 
                           & \% Improvement     & 1.5\%                              \\ \hline
\end{tabular}
\caption{Experiment 1 summary table: n-zone vs 4-nearest zones action spaces}
\label{table:exp1}
\end{table}

\subsubsection{Imbalance algorithm vs Reinforcement Learning algorithms}

\begin{figure}
  \includegraphics[width=\linewidth]{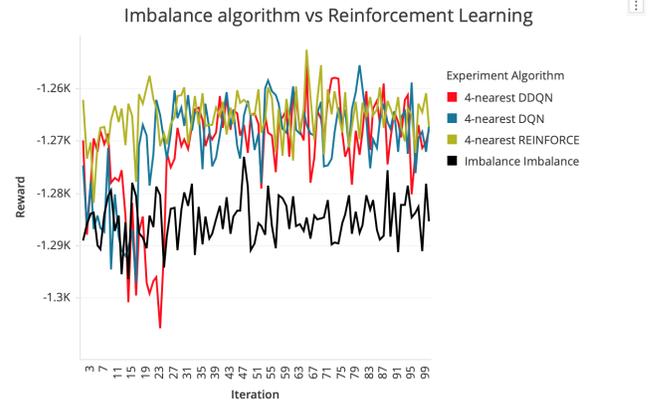}
  \caption{Imbalance algorithm compared with three Reinforcement Learning approaches}
  \label{fig:exp2_imbalance}
\end{figure}

This experiment is to show the benefits of using reinforcement learning algorithms with the 4-nearest zones action space approach compared with the imbalance algorithm. The results clearly indicate that all three reinforcement learning algorithms perform better than \textit{imbalance algorithm}. At the end of the 100th iteration, DQN performs 1.4\% better, DDQN performs 1.4\% better, and REINFORCE performs 1.4\% better compared with \textit{imbalance algorithm} (percentages are rounded up). Figure \ref{fig:exp2_imbalance} shows the comparison in detail. Reasons why \textit{imbalance algorithm} performs worse than  RL algorithms is believed to be because RL algorithms learn from the dynamics of the simulator and they forecast the cities' zones demand and act preemptively to allocate SAV just as the demand spikes. Meanwhile the imbalance algorithm is reactive in nature, it reallocates SAV once there is an imbalance in place, not forecast future imbalances. Then imbalance algorithm is always trying to catch up to the latest situation.

\section{Conclusion}
\label{sec:conclusion}
We studied different reinforcement learning approaches from the basic ideas of Value Iteration and Policy Iteration to the more advanced Monte Carlo Tree Search. Amongst these, the algorithms that support large state space as well as are model free are Deep Q-Learning, Double Deep Q-Learning and REINFORCE. SAV Fleet Management problem has been studied with reinforcement learning approaches before but not with parking cost as a component to optimize. We start from the simulator used by Zhang.W \cite{zhang2018impact} that includes travel times, customer waiting time , and parking costs to investigate whether reinforcement learning can still provide a better optimization compared with existing \textit{imbalance algorithm} \cite{FAGNANT20141}. Simulator was refitted to utilize OpenAI gym APIs. Two experiments were run: which action space performs better?: (1) n-zone action space , each zone has n potential actions where n is the number of zones in the city, and (2) 4-nearest zones action space, each zone has only 4 actions available of the 4 nearest zones. The results of this first experiment is that the 4-nearest zones action space performs better compared with the n-zones approach. This is likely due to easier exploration in an environment with fewer actions. Second experiment investigates whether reinforcement learning approaches perform better than the imbalance algorithm, result is that indeed reinforcement learning  perform better. This is likely due to reinforcement learning algorithms learning to forecast the zones' demand and allocates SAV to zones with most demands as the demand increases in a proactive manner, meanwhile \textit{imbalance algorithm} acts in a reactive manner. Given these consistent, repeatable consistent results, more investigation is needed in same environment with larger service area with larger number of zones. Also, given we are encoding a zone's nearest neighbors , this approach lends itself to graph approaches where nodes can also have nearest neighbors. We can investigate whether framing other NP graph related problems with deep reinforcement learning, such as Travelling Salesman, using the n-nearest nodes action space encoding could provide solutions  in competitive time.


\bibliographystyle{ACM-Reference-Format}
\bibliography{bib}

\end{sloppypar}
\end{document}